\documentclass[10pt,twocolumn,letterpaper]{article}

\usepackage[pagenumbers]{cvpr} 
\usepackage{multirow}
\usepackage{graphicx}
\usepackage{booktabs}
\usepackage{multirow}
\usepackage[table]{xcolor}
\usepackage{colortbl}
\usepackage{array}
\usepackage{float}

\definecolor{cvprblue}{rgb}{0.21,0.49,0.74}
\usepackage[pagebackref,breaklinks,colorlinks,allcolors=cvprblue]{hyperref}

\def\paperID{3} 
\def\confName{CVPR}
\def\confYear{2026}

\title{Scale-Aware Vision-Language Adaptation for Extreme Far-Distance \\ Video Person Re-identification}

\author{Ashwat Rajbhandari \quad Bharatesh Chakravarthi
\\
Arizona State University\\
{\tt\small \{arajbhan, bshettah\}@asu.edu}
}

\usepackage{background}

\newcommand{\mywatermark}{%
    \begin{minipage}{\textwidth}
        \centering
        \fontsize{10}{10}\selectfont 
        This paper has been accepted at the CVPR 2026 Workshop on Human Perception and Recognition in Aerial Surveillance (AERO-HPR)
    \end{minipage}%
}

\backgroundsetup{
    scale=1,
    color=gray,
    angle=0,
    opacity=0.5,
    position=current page.north,
    vshift=-1cm, 
    contents={\mywatermark}
}

\begin{document}
\maketitle
\begin{abstract}
\label{sec:abstract}

Extreme far-distance video person re-identification (ReID) is particularly challenging due to scale compression, resolution degradation, motion blur, and aerial-ground viewpoint mismatch. As camera altitude and subject distance increase, models trained on close-range imagery degrade significantly. 
In this work, we investigate how large-scale vision-language models can be adapted to operate reliably under these conditions. Starting from a CLIP-based baseline, we upgrade the visual backbone from ViT-B/$16$ to ViT-L/$14$ and introduce backbone-aware selective fine-tuning to stabilize adaptation of the larger transformer. To address noisy and low-resolution tracklets, we incorporate a lightweight temporal attention pooling mechanism that suppresses degraded frames and emphasizes informative observations. We retain adapter-based and prompt-conditioned cross-view learning to mitigate aerial-ground domain shifts, and further refine retrieval using improved optimization and k-reciprocal re-ranking. Experiments on the DetReIDX stress-test benchmark show that our approach achieves $46.69$, $41.23$, and $22.98$ mAP on aerial-to-ground (A$2$G), ground-to-aerial (G$2$A), and aerial-to-aerial (A$2$A) protocols, respectively, corresponding to an overall mAP of $35.73$. These results show that large-scale vision-language backbones, when combined with stability-focused adaptation, significantly enhance robustness in extreme far-distance video person ReID.

\end{abstract}    
\section{Introduction}
\label{sec:intro}

\begin{figure}[t]
  \centering
\includegraphics[width=1\linewidth]{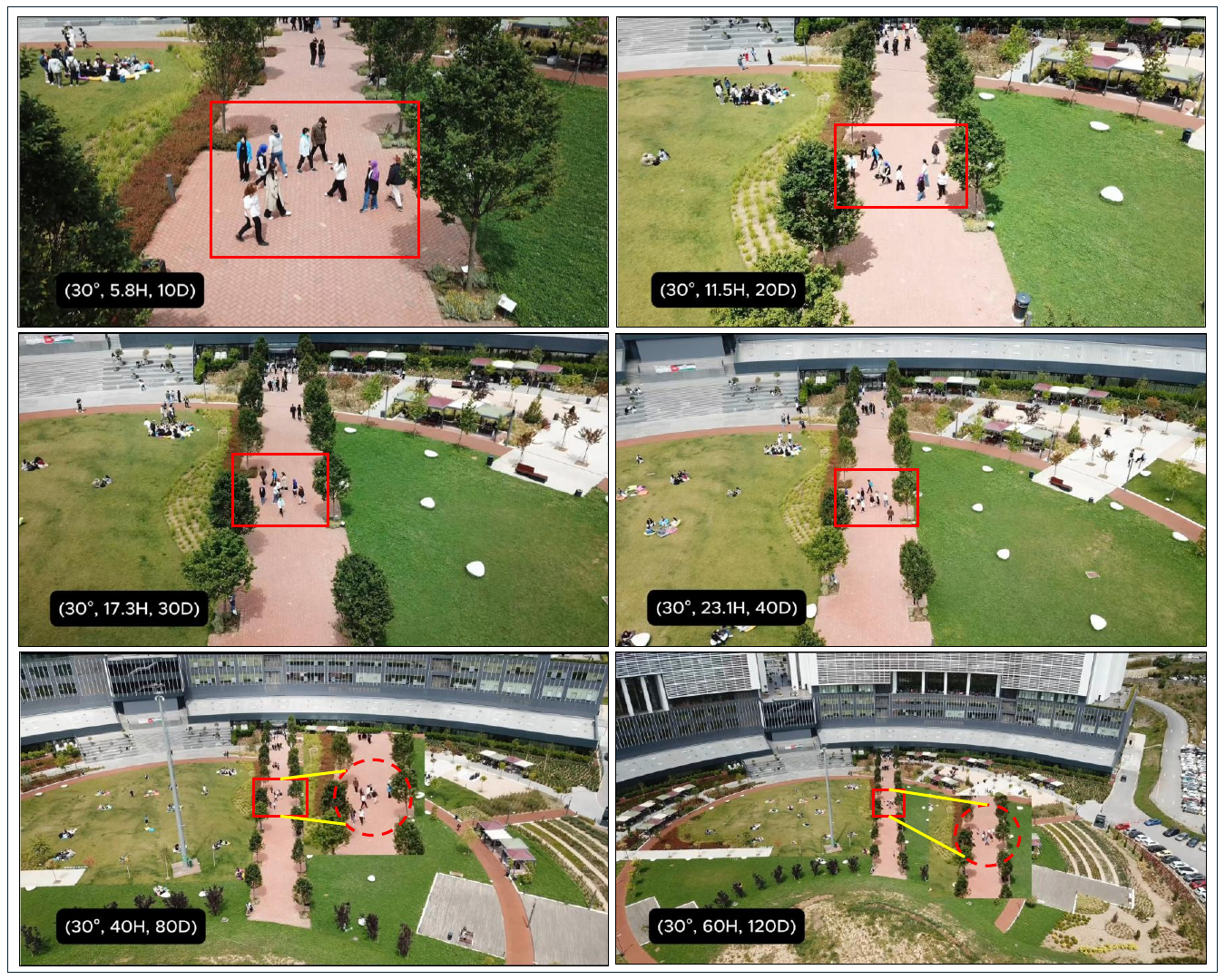}
  \caption{Illustration of extreme far-distance conditions in the DetReIDX dataset. As UAV altitude and horizontal distance increase, pedestrians undergo severe scale compression and resolution degradation. The red bounding boxes highlight the same group of individuals as their pixel footprint shrinks dramatically, showing the visual challenges of aerial-ground video person ReID.}
  \label{fig:fig1}
\end{figure}

Person re-identification (ReID) aims to match individuals across non-overlapping camera views \cite{wu2020reidsurvey} and has progressed rapidly with vision transformers, CLIP-based text encoders, and large-scale pretraining \cite{radford2021clip}. These advances have significantly improved performance on standard ground-level benchmarks and enabled applications in surveillance, crowd analysis, and public safety.
However, extreme far-distance aerial scenarios operate under fundamentally different visual constraints. As the altitude of unmanned aerial vehicles (UAVs) and subject distance increase, pedestrians often occupy only a few pixels and are affected by severe scale compression, resolution degradation, motion blur, and strong viewpoint differences between aerial and ground cameras, as shown in Figure \ref{fig:fig1}. Under such conditions, fine-grained appearance cues become unreliable, and models developed for close-range imagery degrade sharply. The assumptions underlying conventional ReID, namely sufficient spatial detail and consistent viewpoints, no longer apply.

Recent stress-test datasets and benchmarks for UAV-based person analysis highlight this limitation explicitly. In particular, the DetReIDX \cite{hambarde2025detreidx} was introduced to evaluate robustness under long-range viewpoints, aerial-ground cross-domain matching, and session-level appearance changes. The benchmark reveals that strong detection and ReID baselines can collapse under extreme distance and viewpoint variation. Improving performance, therefore, requires adaptation strategies that explicitly account for scale degradation and noisy video tracklets, rather than direct transfer from conventional settings.

In this work, we explore how large-scale vision-language backbones can be effectively adapted for extreme far-distance video ReID. Starting from the official CLIP ViT-B/$16$ baseline for DetReIDX, we investigate backbone scaling and stability-oriented fine-tuning. We upgrade the visual encoder to ViT-L/$14$ and introduce backbone-aware selective fine-tuning to preserve pretrained representations while enabling high-level domain adaptation. To improve robustness to frame-level degradation, we incorporate a lightweight temporal attention pooling mechanism that emphasizes informative frames within noisy tracklets. We further retain prompt-based cross-view conditioning and adapter tuning to mitigate aerial-ground domain shifts, and refine retrieval performance through optimized training strategies and k-reciprocal re-ranking \cite{zhong2017reranking}. 
On the DetReIDX benchmark, evaluated across aerial-to-aerial (A$2$A), aerial-to-ground (A$2$G), and ground-to-aerial (G$2$A) protocols \cite{hambarde2025detreidx}, our approach improves query weighted overall mAP from $28.11$ to $35.73$ and achieves strong gains across all evaluation protocols, including $46.69$ on A$2$G, $41.23$ on G$2$A, and $22.98$ on A$2$A, demonstrating consistent improvements under both cross-view and same-view extreme far-distance conditions. These findings demonstrate that large-scale CLIP backbones, when carefully adapted, substantially improve robustness in extreme far-distance video person ReID.
We summarize the contribution of this work below:

\begin{enumerate}
    \item We present a systematic study of CLIP backbone scaling (ViT-B/$16$ to ViT-L/$14$) for extreme far-distance aerial-ground video person ReID, demonstrating that increased model capacity improves robustness under severe scale compression.  
    \item We propose a stability-focused adaptation strategy combining backbone-aware selective fine-tuning, temporal attention pooling, and prompt/adapter tuning to handle noisy, low-resolution video tracklets.
    \item We introduce practical optimization refinements, including cosine scheduling and enhanced data augmentation, and enable k-reciprocal re-ranking at inference.
    \item Our method achieves $35.73$ mAP on the benchmark, improving substantially from the baseline ($28.11$) and the highest publicly reported score ($32.89$). 
\end{enumerate}

The remainder of this paper is organized as follows. Section \ref{sec:relatedWork} reviews related work in ground-level and aerial video person ReID, vision-language adaptation, and retrieval re-ranking. Section \ref{sec:methodology} describes the proposed scale-aware adaptation framework, including backbone scaling, selective fine-tuning, temporal attention pooling, and optimization refinements. Section \ref{sec:experiments} presents experimental results on the DetReIDX benchmark, including implementation details and quantitative comparisons. Section \ref{sec:discussion} discusses the results, limitations, and future improvements. Finally, Section \ref{sec:conclusion} concludes the paper.

\section{Related Work}
\label{sec:relatedWork}

Person ReID has been extensively studied in both image-based and video-based settings. With the adoption of transformer backbones and large-scale pretraining, significant progress has been achieved on conventional ground-level benchmarks. However, robustness under extreme far-distance aerial conditions remains largely underexplored. We review related work in ground-level ReID, video-based temporal modeling, aerial and cross-view benchmarks, vision-language adaptation, and retrieval re-ranking.  

\subsection{Ground-level Person ReID}
Early ReID research focused primarily on ground-level camera networks, where pedestrians are captured at moderate distances with sufficient spatial resolution. Benchmarks such as Market-1501 \cite{zheng2015market} and DukeMTMC4ReID \cite{ristani2016dukemtmc,zheng2017unlabeled} established standardized evaluation protocols and drove rapid development of discriminative feature learning methods. More recently, transformer-based architectures, including TransReID \cite{he2021transreid} and related global-context modeling frameworks, have demonstrated strong performance by leveraging self-attention and large-scale pretraining.
Despite these advances, most methods are evaluated under relatively stable imaging conditions. When applied to extreme far-distance aerial scenarios, where person instances are severely scale-compressed and contain limited visual detail, their performance degrades substantially. This limitation underscores the challenge of directly transferring ground-level transformer models to aerial settings without explicit adaptation to scale degradation and viewpoint discrepancies.

\begin{table*}[t]
\centering
\caption{Comparison of aerial-ground benchmark datasets for person detection, ReID, tracking, and action recognition.}
\label{tab:dataset_comparison}
\small
\begin{tabular}{l c c c c c c}
\toprule
\textbf{Dataset} & \textbf{Camera} & \textbf{Format} & \textbf{PIDs} & \textbf{BBoxes} & \textbf{Height (m)} & \textbf{Distance (m)} \\
\midrule
AG-ReID.v2 & UAV+CCTV & Still & 1615 & 100.6K & 15-45 & – \\
G2APS-ReID & UAV+CCTV & Still & 2788 & 200.8K & 20-60 & – \\
DetReIDX & DSLR+UAV & Video+Still & 334 & 13M & 5-120 & 10-120 \\
\bottomrule
\end{tabular}
\end{table*}

\subsection{Video-based ReID and Temporal Modeling}

Video-based ReID extends image-based matching by exploiting temporal cues across tracklets. The MARS benchmark introduced large-scale video evaluation and motivated sequence aggregation strategies beyond simple frame averaging \cite{zheng2016mars}. Early approaches employed recurrent architectures or heuristic pooling, while more recent works adopt attention-based mechanisms to emphasize informative frames and suppress noise. Methods such as spatio-temporal attention (STA) \cite{fu2019sta} and temporal complementary learning networks (TCLNet) \cite{hou2020tclnet} demonstrate the effectiveness of adaptive aggregation for handling occlusion, pose variation, and motion blur. In extreme far-distance aerial footage, tracklets often contain heavily degraded frames due to motion instability and severe resolution loss. Under such conditions, uniform aggregation can amplify noise, making adaptive temporal weighting particularly important.

\subsection{Aerial and Cross-View ReID Benchmarks}
UAV-based person analysis introduces challenges that are not captured by conventional ground-only datasets. UAV-Human \cite{li2021uavhuman} and P-DESTRE \cite{kumar2021pdestre} extend ReID to aerial platforms, incorporating detection, tracking, and both short and long-term matching tasks. For explicit aerial-ground cross-view matching, AG-ReID \cite{nguyen2023aerialground} and related datasets reveal a substantial appearance gap between top-down UAV imagery and horizontal ground-camera imagery. More recently, DetReIDX (see Table~\ref{tab:dataset_comparison})  was proposed as a stress-test benchmark \cite{hambarde2025detreidx} targeting extreme far-distance conditions. It explicitly models altitude variation, scale compression, cross-view mismatch, and session-level appearance drift. Experimental findings show that strong detection and ReID baselines can collapse under such conditions, underscoring the need for scale-aware and stability-oriented adaptation strategies. These characteristics make DetReIDX an appropriate benchmark for evaluating robustness in extreme far-distance aerial-ground ReID.

\subsection{Vision-Language Pretraining and Efficient Adaptation}
Vision-language models (VLMs), particularly CLIP, have demonstrated strong transferability through large-scale image-text pretraining \cite{radford2021clip}. CLIP-based features have been incorporated into ReID frameworks to improve generalization and mitigate the absence of semantic class labels. Approaches such as CLIP-ReID leverage prompt learning to bridge identity supervision and semantic embedding spaces. Beyond prompt optimization, parameter-efficient adaptation methods, including adapters, low-rank updates, and conditional prompts (e.g., CoOp and CoCoOp) \cite{zhou2022cocoop}, enable stable fine-tuning of large pretrained backbones while reducing overfitting risk. Such strategies are particularly relevant in extreme far-distance aerial-ground ReID, where severe degradation and limited effective visual detail increase the risk of representation drift during adaptation.

\subsection{Re-Ranking for Retrieval-based ReID}
Post-processing remains an important component of retrieval-based ReID systems. In particular, k-reciprocal re-ranking refines similarity relationships by exploiting reciprocal nearest neighbors in the embedding space, leading to consistent improvements in mAP and ranking accuracy. Careful parameter tuning is often required to maximize gains for a given backbone and feature distribution \cite{zhong2017reranking}.
In contrast to prior work, which primarily focuses on ground-level settings or moderate aerial conditions, we target the extreme far-distance regime evaluated by DetReIDX. Such recent benchmark evaluations have shown that performance degradation is particularly pronounced in cross-view settings such as A$2$G and G$2$A retrieval, where viewpoint mismatch and scale variation are most severe. This highlights the need for adaptation strategies that not only improve overall performance but also address protocol-specific challenges. Hence, we investigate how large-scale vision-language backbones can be systematically scaled and stably adapted to address severe scale compression and noisy video tracklets. Section \ref{sec:methodology} describes our proposed scale-aware adaptation framework in detail. 
\section{Scale-Aware Adaptation Framework}
\label{sec:methodology}

We build upon the official CLIP-based \cite{radford2021clip} video ReID baseline for the DetReIDX benchmark and introduce a series of scale-aware modifications tailored for extreme far-distance (XFD) aerial-ground scenarios. Our approach focuses on three key aspects: ($1$) increasing backbone capacity through model scaling, ($2$) stabilizing adaptation of the larger transformer, and ($3$) improving robustness to degraded video tracklets via adaptive temporal aggregation and refined optimization. We first describe the baseline framework and then present the proposed modifications.

\subsection{Baseline Framework}
\label{sec:baseline}

We adopt the official CLIP-based video ReID baseline released for DetReIDX. The framework performs tracklet-level retrieval under aerial-ground cross-view conditions and includes a vision transformer backbone, prompt-based cross-view conditioning, and parameter-efficient adaptation modules.
\paragraph{ \textit{Backbone and frame-level feature extraction.}}
The baseline employs a Vision Transformer backbone with patch size $16\times16$ (ViT-B/16). All frames are resized to $256\times128$ during both training and testing \cite{dosovitskiy2021vit}.

\paragraph{\textit{Tracklet construction and sampling.}}
Video ReID is performed at the tracklet level. For each training instance, a fixed-length sequence of $L=16$ frames is sampled and encoded independently by the backbone. A softmax triplet sampling strategy is adopted to support both identity classification and metric learning \cite{hermans2017triplet}. The batch size is set to $16$ tracklets. During inference, the same sequence length is used, and retrieval is conducted using cosine similarity between $\ell_2$-normalized features.

\paragraph{\textit{Prompt-based cross-view conditioning (PBP).}}
To reduce aerial-ground domain mismatch, the baseline incorporates prompt-based cross-view conditioning. A prompt of length $1$ is inserted across $9$ transformer layers. Metadata signals, including altitude, horizontal distance, and viewing angle, are discretized into bins ($18$ altitude, $18$ distance, and $3$ angle bins) and encoded as conditioning inputs. This design injects physical context related to viewpoint and scale variation into the representation learning process.

\paragraph{\textit{Training strategy.}}

The baseline is trained using a multi-term objective combining identity classification, triplet metric learning, and CLIP-style cross-modal alignment. The overall loss is defined as:

\begin{equation}
\mathcal{L} = \lambda_{\text{id}}\mathcal{L}_{\text{id}} 
+ \lambda_{\text{tri}}\mathcal{L}_{\text{tri}} 
+ \lambda_{\text{i2t}}\mathcal{L}_{\text{i2t}} 
+ \lambda_{\text{t2i}}\mathcal{L}_{\text{t2i}},
\end{equation}

where the loss weights and optimization parameters are summarized in Table~\ref{tab:training_config}.

\begin{table}[t]
\centering
\caption{Baseline training configuration.}
\label{tab:training_config}
\footnotesize
\setlength{\tabcolsep}{4pt}
\renewcommand{\arraystretch}{1.05}
\begin{tabular}{l c c}
\toprule
\textbf{Parameter} & \textbf{Stage 1} & \textbf{Stage 2} \\
\midrule
Optimizer & Adam & Adam \\
Base Learning Rate & $3.5 \times 10^{-4}$ & $1 \times 10^{-4}$ \\
Max Epochs & 120 & 120 \\
Weight Decay & $1 \times 10^{-4}$ & $2.5 \times 10^{-4}$ \\
Weight Decay (Bias) & $1 \times 10^{-4}$ & $1 \times 10^{-4}$ \\
Images Per Batch & 16 & 16 \\
\midrule
ID Loss Weight ($\lambda_{\text{id}}$) & \multicolumn{2}{c}{$0.25$} \\
Triplet Loss Weight ($\lambda_{\text{tri}}$) & \multicolumn{2}{c}{$1.0$} \\
I2T / T2I Weights & \multicolumn{2}{c}{$1.0$} \\
\bottomrule
\end{tabular}
\end{table}

Stage $1$ of training focuses on stable representation learning with cosine-style decay, with some warm-up. Stage $2$ training performs further fine-tuning using a reduced learning rate and multi-step scheduling to refine retrieval performance and improve ranking stability.

\paragraph{\textit{Inference settings.}}
During evaluation, tracklet features are extracted using fixed-length sequences of $16$ frames and $\ell_2$ normalized before retrieval. Cosine similarity is used for distance computation. Re-ranking is disabled in the default baseline configuration, considering it may degrade the learned representation. The full inference configuration is summarized in Table~\ref{tab:inference_config}.

\begin{table}[t]
\centering
\caption{Baseline inference configuration.}
\label{tab:inference_config}
\footnotesize
\setlength{\tabcolsep}{5pt}
\renewcommand{\arraystretch}{1.05}
\begin{tabular}{l c}
\toprule
\textbf{Parameter} & \textbf{Setting} \\
\midrule
Batch Size & 16 \\
Sequence Length & 16 \\
Feature Normalization & $\ell_2$ normalization \\
Distance Metric & Cosine similarity \\
Re-ranking & Disabled \\
\bottomrule
\end{tabular}
\end{table}

\paragraph{\textit{Parameter-efficient adaptation via adapters.}}
The baseline further uses lightweight adapter modules \cite{houlsby2019adapters} as a parameter-efficient mechanism for adapting the pretrained transformer to the target domain. Other optional components provided by the baseline, such as VCAH and QATW, are disabled initially. Temporal Attention pooling is also disabled in the baseline, resulting in the use of a standard mean pooling.

\subsection{Scale-Aware Adaptation}

\begin{figure}[ht]
  \centering
\includegraphics[width=1\linewidth]{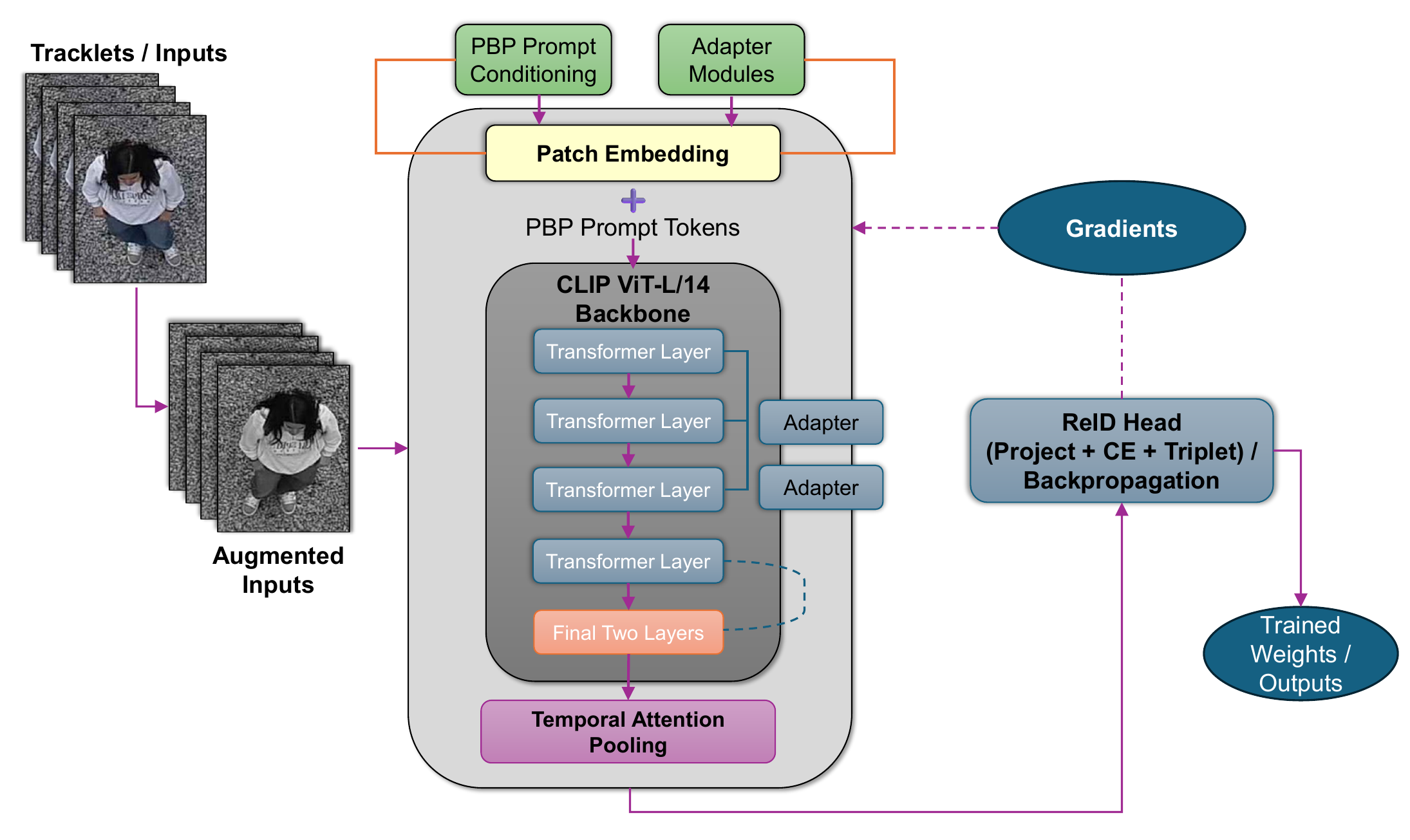}
  \caption{Overview of the proposed CLIP ViT-L/$14$-based video ReID framework. Augmented tracklets are encoded by a CLIP backbone enhanced with PBP prompt conditioning and adapter modules, with only the final two transformer blocks unfrozen during fine-tuning. Frame-level features are aggregated via temporal attention pooling and optimized using identity and triplet losses.}
  \label{fig:fig2}
\end{figure}

\label{sec:mods}
While the baseline provides a strong foundation for extreme far-distance video ReID, performance remains limited under severe scale compression and heavy frame degradation. We therefore introduce several targeted modifications focused on backbone capacity, stable large-model adaptation, robust temporal aggregation, re-ranking, and optimized training strategies. An overview of the proposed pipeline is shown in Figure \ref{fig:fig2}.

\subsubsection{Backbone Scaling: ViT-B/$16$ $\rightarrow$ ViT-L/$14$}
\label{sec:mods_backbone}

We modify the baseline configuration by switching the CLIP visual encoder from ViT-B/$16$ to the larger ViT-L/$14$ backbone, thereby increasing representational capacity under severe resolution degradation and scale compression. In very far-distance settings, pedestrian regions often occupy only a small number of pixels, and fine-grained appearance cues, such as texture and local structure, become unreliable. A larger transformer backbone provides improved modeling capacity to extract robust global and mid-level features from such degraded inputs \cite{radford2021clip,dosovitskiy2021vit}.
ViT-B/16 produces $768$-dimensional embeddings, whereas ViT-L/$14$ outputs $1024$-dimensional embeddings. Consequently, we adjust the associated projection layers, classification heads, prompt embeddings, and adapter modules to ensure dimensional consistency throughout the architecture. These adjustments are strictly architectural and do not alter the training objective or data sampling strategy. Apart from the backbone upgrade and the necessary dimensional alignment, the overall training pipeline is preserved. This design allows us to isolate the impact of backbone scaling while maintaining comparability with the original baseline.

\subsubsection{Backbone-Aware Selective Fine-Tuning}
\label{sec:mods_ft}
Fine-tuning all parameters of a large transformer can degrade the strong feature representations learned during large-scale pretraining and lead to unstable optimization and overfitting under extreme far-distance conditions. Hence, instead of training the entire architecture on this dataset, the baseline had most backbone parameters frozen, focusing on adapter and prompt training, while we unfroze the high-level blocks (i.e., blocks $22$ and $23$) for adaptation. The final two blocks were unfrozen since the earlier layers encode general visual features. This strategy preserves the pretrained model’s generalization ability while enabling high-level representations to adapt to the target domain \cite{houlsby2019adapters,hu2022lora}.
We further use differential learning rates by assigning a smaller learning rate to the unfrozen backbone blocks and a larger learning rate to lightweight modules and heads (e.g., prompts/adapters and the identity classifier). Specifically, the unfrozen backbone blocks are trained using a learning rate scaled to $0.1$× of the base learning rate. This improves training stability when adapting the larger ViT-L/$14$ model and helps avoid catastrophic drift from pretrained weights.

\subsubsection{Temporal Attention Pooling}
\label{sec:mods_temporal}

The baseline uses a standard mean pooling because attention pooling is disabled. However, in extreme far-distance aerial settings, individual frames often suffer from severe blur, motion artifacts, occlusion, or extreme scale compression. Standard mean pooling can, therefore, allow degraded frames to negatively influence the final representation. To address this limitation, we enable and introduce a lightweight temporal attention mechanism that adaptively weights frames based on their feature responses \cite{fu2019sta,hou2020tclnet}. Let $\{f_t\}_{t=1}^{T}$ denote the sequence of frame-level embeddings extracted from a tracklet, where $T=16$ in our setting and $f_t \in \mathbb{R}^{C}$.

We learn a scalar attention score for each frame using a linear projection:
\begin{equation}
s_t = w^\top f_t,
\end{equation}
where $w \in \mathbb{R}^{C}$ is a learnable parameter vector implemented as a fully connected layer. The attention weights are obtained via softmax normalization across the temporal dimension:
\begin{equation}
\alpha_t = \frac{\exp(s_t)}{\sum_{k=1}^{T} \exp(s_k)}.
\end{equation}

The final tracklet representation is computed as a weighted sum:
\begin{equation}
z = \sum_{t=1}^{T} \alpha_t f_t.
\end{equation}

This mechanism enables the model to suppress degraded or irrelevant frames while emphasizing temporally consistent and discriminative observations. In our implementation, temporal attention is applied consistently to the intermediate and projected feature representations before classification and retrieval. Compared to the baseline, this adaptive pooling strategy improves robustness under extreme resolution degradation and unstable aerial motion.

\begin{table}[t]
\centering
\caption{Optimization refinements compared to the baseline configuration.}
\label{tab:opt_refinement}
\footnotesize
\setlength{\tabcolsep}{4pt}
\renewcommand{\arraystretch}{0.95}
\begin{tabular}{l c c}
\toprule
\textbf{Parameter} & \textbf{Baseline (ViT-B/16)} & \textbf{Ours (ViT-L/14)} \\
\midrule
Backbone & ViT-B/16 & ViT-L/14 \\
Stride Size & 16 & 14 \\
PBP Prompt Length & 1 & 4 \\
QATW Module & Disabled & Enabled \\
Instance Norm & Disabled & Enabled \\
\midrule
\multicolumn{3}{c}{\textbf{Stage 1 Training}} \\
\midrule
Optimizer & Adam & Adam \\
Images Per Batch & 16 & 48 \\
Base Learning Rate & $3.5 \times 10^{-4}$ & $2.0 \times 10^{-4}$ \\
Max Epochs & 120 & 50 \\
LR Schedule & Cosine Annealing & Cosine Annealing \\
Weight Decay & $1 \times 10^{-4}$ & $1 \times 10^{-4}$ \\
\midrule
\multicolumn{3}{c}{\textbf{Stage 2 Fine-Tuning}} \\
\midrule
Optimizer & Adam & Adam \\
Images Per Batch & 16 & 24 \\
Base Learning Rate & $1 \times 10^{-4}$ & $1 \times 10^{-4}$ \\
Max Epochs & 120 & 40 \\
LR Schedule & Multi-step (60, 90) & Cosine Annealing \\
\bottomrule
\end{tabular}
\end{table}

\subsubsection{Optimization Strategy}
\label{sec:mods_optimization}

In addition to backbone scaling and selective fine-tuning, we refine the optimization strategy to better accommodate the larger ViT-L/$14$ backbone and improve training stability under extreme far-distance conditions. Compared to the baseline configuration, we adjust batch size, learning rates, training duration, and scheduling strategy to balance convergence speed and generalization.
Table~\ref{tab:opt_refinement} summarizes the key differences between the baseline and our refined optimization setup. Specifically, we increase the Stage~$1$ batch size from $16$ to $48$ to expose the larger backbone to more identity diversity per iteration. The base learning rate for Stage~$1$ is slightly reduced (from $3.5\times10^{-4}$ to $2.0\times10^{-4}$) to stabilize optimization when training the deeper ViT-L/$14$ architecture. The total number of training epochs is also reduced in both stages (Stage~$1$: $120$ $\rightarrow$ $50$, Stage~$2$: $120$ $\rightarrow$ $40$), reflecting faster convergence with the scaled backbone.

In Stage~$1$, the selectively unfrozen backbone blocks are trained with a reduced learning rate scaled to $0.1\times$ the base learning rate, as described in Section~\ref{sec:mods_ft}, to prevent destabilization of pretrained representations. 
Unlike the baseline, which uses multi-step decay during Stage~$2$, our refined configuration adopts cosine annealing \cite{loshchilov2017sgdr} for both training stages. This smooth decay schedule avoids abrupt learning rate drops and provides more stable adaptation of the selectively unfrozen transformer blocks. Additionally, instance normalization at the neck is enabled to further improve robustness under extreme scale degradation. These refinements collectively stabilize large-model adaptation and improve convergence behavior under severe resolution loss.

\subsubsection{Data Augmentation}
\label{sec:mods_augmentation}

To improve robustness under diverse aerial capture conditions, we extend the baseline augmentation pipeline by incorporating color jitter during training. Specifically, we apply controlled perturbations in brightness, contrast, saturation, and hue to simulate illumination variation and camera-induced color shifts commonly observed in extreme far-distance imagery. 
In addition to random horizontal flipping and random erasing used in the baseline, these color augmentations encourage the model to rely less on fragile color cues and more on structural and identity-consistent features.
During inference, we further apply horizontal flip augmentation and average the features extracted from the original and flipped sequences to improve orientation robustness.

\subsubsection{k-Reciprocal Re-Ranking}
\label{sec:mods_rerank}

To further refine retrieval results, we enable k-reciprocal re-ranking as a post-processing step during inference. After computing cosine distances between $\ell_2$-normalized query and gallery embeddings, the initial distance matrix is refined using k-reciprocal encoding to improve neighborhood consistency in the embedding space.
Specifically, we apply re-ranking with parameters $(k_1 = 28,\, k_2 = 6,\, \lambda = 0.28)$. This procedure re-evaluates similarity relationships by considering reciprocal nearest neighbors, reducing the impact of local feature noise, and improving retrieval robustness.
Re-ranking is applied only during evaluation and does not affect model training. This inference refinement yields additional improvements in the mean Average Precision (mAP) score.

\begin{table*}[t]
\centering
\footnotesize
\setlength{\tabcolsep}{3.5pt}
\renewcommand{\arraystretch}{1.1}
\caption{Comparison of baseline methods and our method on VReID-XFD.}
\label{tab:table5}
\begin{tabular}{l|cccc|cccc|cccc}
\toprule
\multirow{2}{*}{\textbf{Method}} 
& \multicolumn{4}{c|}{\textbf{A$\rightarrow$A}} 
& \multicolumn{4}{c|}{\textbf{A$\rightarrow$G}} 
& \multicolumn{4}{c}{\textbf{G$\rightarrow$A}} \\
\cmidrule(lr){2-5} \cmidrule(lr){6-9} \cmidrule(lr){10-13}
& R1 & R5 & R10 & mAP & R1 & R5 & R10 & mAP & R1 & R5 & R10 & mAP \\
\midrule
VSLA~\cite{zhang2024crossplatform}       
& 15.96 & 26.10 & 32.77 & 13.83 
& 28.96 & 54.71 & 69.13 & 41.63 
& 58.43 & 65.17 & 69.66 & 26.26 \\

SINet~\cite{bai2022salient}         
& 14.06 & 24.51 & 30.94 & 12.85 
& 25.62 & 52.47 & 66.57 & 38.46 
& 23.50 & 49.44 & 59.55 & 16.98 \\

PSTA~\cite{wang2021pyramid}         
& 13.00 & 23.80 & 30.30 & 10.50 
& 22.30 & 46.90 & 59.70 & 34.40 
& 40.40 & 56.20 & 59.60 & 17.00 \\

BiCNet-TKS~\cite{hou2021bicnet}  
& 13.30 & 26.78 & 36.57 & 9.71 
& 21.71 & 44.85 & 59.30 & 33.28 
& 41.57 & 58.43 & 65.17 & 22.12 \\

\midrule
DUT\_IIAU\_LAB~\cite{hambarde2026vreidxfd}  
& 25.39 & 39.58 & 48.44 & 20.13 
& 37.77 & 65.26 & 75.31 & 43.93 
& 69.66 & 80.90 & 87.64 & 35.44 \\

\midrule
\textbf{Ours (CLIP ViT-L/14)}   
& 22.70 & 30.91 & 35.75 & \textbf{22.98} 
& 41.42 & 61.95 & 74.16 & \textbf{46.69} 
& 70.79 & 71.91 & 74.16 & \textbf{41.23} \\
\bottomrule
\end{tabular}
\end{table*}

\section{Experiment \& Results}
\label{sec:experiments}

The experiments were conducted on a high-performance computing (HPC) cluster using NVIDIA A$100$ GPUs. Each training run utilized $8$ CPU cores and $48$\ GiB of system memory. This setup was sufficient to train the scaled ViT-L/$14$ backbone under the proposed adaptation strategy.

\subsection{DetReIDX Dataset}
We train and evaluate our model on the DetReIDX stress-test benchmark \cite{hambarde2025detreidx}, which is designed to evaluate aerial surveillance person ReID under extreme far-distance degradation and aerial-ground cross-view mismatch. The final benchmark score is computed as the mean Average Precision (mAP) over all query instances across the three protocols (A$2$G, G$2$A, and A$2$A). Since the number of queries differs significantly across protocols, the overall mAP reflects a query-weighted aggregation rather than a simple average across domains. Formally, the overall score is computed as:

\begin{equation}
\resizebox{\columnwidth}{!}{$
\text{mAP}_{\text{overall}} = 
\frac{N_{\text{A2G}} \cdot \text{mAP}_{\text{A2G}} + 
      N_{\text{G2A}} \cdot \text{mAP}_{\text{G2A}} + 
      N_{\text{A2A}} \cdot \text{mAP}_{\text{A2A}}}
     {N_{\text{A2G}} + N_{\text{G2A}} + N_{\text{A2A}}}
$}
\end{equation}

where $N_{\text{A2G}}$, $N_{\text{G2A}}$, and $N_{\text{A2A}}$ denote the number of query instances in each protocol.

We use mean average precision (mAP) as the primary evaluation metric, following standard ReID protocols. Due to the imbalance in query distribution, the overall score is primarily influenced by performance on the A$2$G and A$2$A protocols, while G$2$A contributes comparatively less.

\subsection{Implementation Details}

\paragraph{\textit{Backbone and Training Setup.}}
The visual encoder is initialized from the CLIP ViT-L/$14$ checkpoint pretrained on large-scale image–text data. Training is conducted in two stages. In Stage~$1$, the model is trained for $50$ epochs with a batch size of $48$ tracklets and a base learning rate of $2\times10^{-4}$. Stage~$2$ performs fine-tuning for $40$ epochs using a batch size of $24$ and a base learning rate of $1\times10^{-4}$.
Selective fine-tuning is applied by unfreezing only the final two transformer blocks (resblocks.22 and resblocks.23), while earlier layers remain frozen. The unfrozen blocks are optimized with a learning rate scaled to $0.1\times$ the base rate to preserve pretrained representations and stabilize adaptation. Adam is used as the optimizer in both training stages.

\paragraph{\textit{Prompt and Conditioning Configuration.}}

Prompt-based cross-view conditioning is enabled in the final model with a prompt length of $4$ and deep prompt insertion across $9$ transformer layers. Camera and metadata conditioning, based on discretized altitude, distance, and viewing angle bins, are activated as described in Section~\ref{sec:baseline}. Adapter modules, the QATW module, and instance normalization are also enabled in the final configuration.

\paragraph{\textit{Learning Rate Scheduling and Regularization.}}
Cosine annealing is used in both training stages to provide smooth learning rate decay and stable convergence. Weight decay is set to $1\times10^{-4}$ in Stage~$1$ and $2.5\times10^{-4}$ in Stage~$2$. 

\paragraph{\textit{Data Processing and Augmentation.}}

All frames are resized to $256\times128$ and normalized. Each tracklet consists of $16$ sampled frames. During training, we apply random horizontal flipping ($p=0.5$), random erasing ($p=0.5$), color jitter with brightness, contrast, saturation, and hue parameters of $(0.1, 0.1, 0.1, 0.05)$, and padding of $10$ pixels. These augmentations improve robustness to noise and enhance generalization under aerial degradation conditions.

\paragraph{\textit{Temporal Aggregation and Inference.}}
Temporal attention pooling aggregates frame-level embeddings into tracklet-level representations. A lightweight linear attention module computes adaptive frame weights through softmax normalization along the temporal dimension, enabling the model to emphasize informative frames while suppressing degraded ones. During inference, tracklet embeddings are $\ell_2$-normalized, and cosine similarity is used for retrieval.

\paragraph{\textit{K-reciprocal re-ranking.}}

To further improve retrieval consistency, we apply k-reciprocal re-ranking as a post-processing step during inference. After computing cosine distances between $\ell_2$-normalized query and gallery embeddings, the distance matrix is refined using k-reciprocal encoding. Unless otherwise specified, we set $k_1=28$, $k_2=6$, and $\lambda=0.28$. Re-ranking is applied only during evaluation and does not affect model training.

\begin{figure}[ht]
  \centering
\includegraphics[width=1\linewidth]{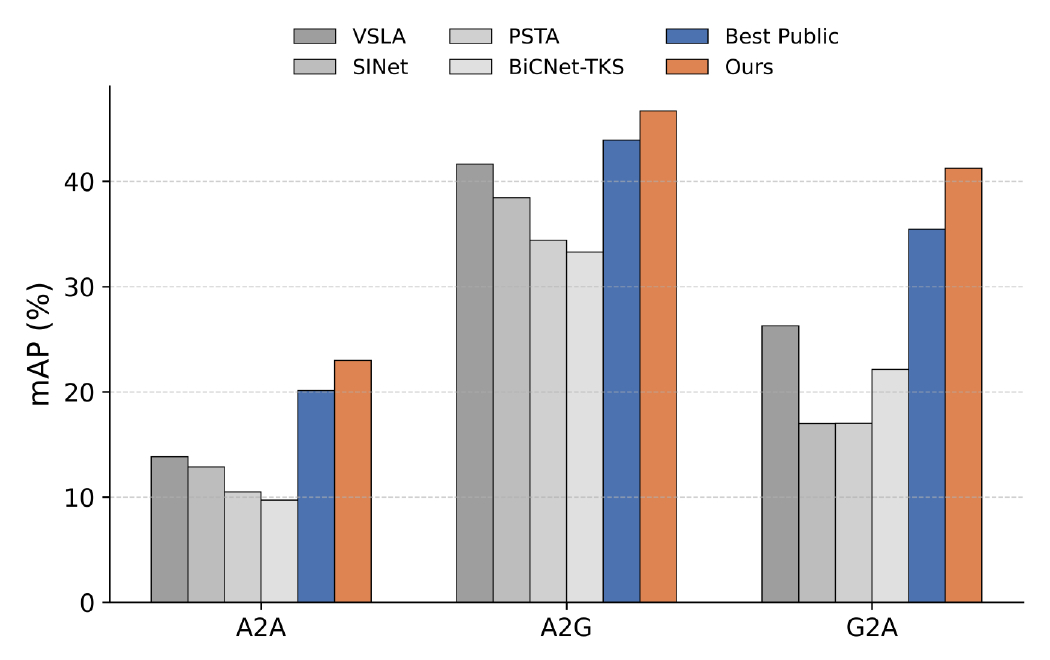}
    \caption{mAP comparison across baseline methods, the best public result, and our method on the DetReIDX benchmark.}
  \label{fig:fig4}
\end{figure}

\subsection{Results}
\label{sec:results}

We report results on DetReIDX using the official evaluation protocol defined in Eq.~(5), where the overall mAP is computed over all query instances across the three protocols (A$2$G, G$2$A, and A$2$A).
As shown in Table \ref{tab:table5} and Figure \ref{fig:fig4}, our full model achieves an overall mAP of $35.73$, improving over the official CLIP ViT-B/$16$ baseline by +$7.62$ and over the best public result by +$2.84$. Protocol-wise, the proposed method attains $46.69$ on A$2$G, $41.23$ on G$2$A, and $22.98$ on A$2$A, outperforming the best public result across all three settings. The largest gain is observed on G$2$A, indicating that the proposed stability-aware adaptation is particularly effective under severe cross-view mismatch and scale degradation.
Beyond mAP, our method also improves Rank-$1$ in the two cross-view protocols, while gains in Rank-$5$ and Rank-$10$ are comparatively smaller. This suggests that the proposed approach primarily strengthens top-ranked retrieval and overall ranking quality rather than uniformly improving deeper positions in the ranked list.

\subsection{Ablation Study}
\label{sec:ablation}

We perform ablation experiments to quantify the contribution of each component in the proposed scale-aware adaptation framework. All results are reported using the official overall mAP defined in Eq.~(5). As shown in Table \ref{tab:table6}, the official CLIP ViT-B/16 baseline achieves $28.11$ overall mAP. Applying optimization refinements yields a modest gain to $28.18$, while enabling k-reciprocal re-ranking improves performance to $29.40$. Further tuning of the re-ranking configuration increases the score to $29.47$, and incorporating temporal attention pooling raises performance to $29.71$. The final ViT-B/16 configuration reaches $29.84$, corresponding to a total improvement of +$1.73$ over the baseline.
The largest gain comes from backbone scaling and stability-aware adaptation. Replacing ViT-B/16 with ViT-L/14 and selectively adapting the larger backbone boosts overall mAP to $35.73$, yielding a substantial +$7.62$ improvement over the baseline. These results indicate that re-ranking and temporal attention provide consistent complementary benefits within the smaller backbone, while increased representational capacity becomes most effective when coupled with stable adaptation under extreme far-distance degradation.

\begin{table}[t]
\centering
\caption{Ablation study on DetReIDX using Eq.~5.}
\label{tab:table6}
\footnotesize
\setlength{\tabcolsep}{3.5pt}
\renewcommand{\arraystretch}{1.05}
\resizebox{\columnwidth}{!}{
\begin{tabular}{l|l|c|c}
\toprule
\multicolumn{1}{c|}{\textbf{Variant}} &
\multicolumn{1}{c|}{\textbf{Backbone}} &
\multicolumn{1}{c|}{\textbf{mAP (\%)}} &
\multicolumn{1}{c}{\textbf{$\Delta$}} \\
\midrule
CLIP ViT-B/16                   & \multirow{6}{*}{ViT-B/16} & 28.11 & +0.00 \\
+ Optimization                  &                           & 28.18 & +0.07 \\
+ k-reciprocal re-ranking       &                           & 29.40 & +1.29 \\
+ Re-ranking tuned              &                           & 29.47 & +1.36 \\
+ Temporal attention pooling    &                           & 29.71 & +1.60 \\
+ Re-ranking tuned (final)      &                           & 29.84 & +1.73 \\
\midrule
+ Backbone scaling + Adaptation & ViT-L/14                  & \textbf{35.73} & \textbf{+7.62} \\
\bottomrule
\end{tabular}
}
\end{table}

\section{Discussion}
\label{sec:discussion}

Our results show that extreme far-distance video person ReID benefits primarily from increased representational capacity when adaptation remains stable. Scaling the CLIP visual encoder from ViT-B/$16$ to ViT-L/$14$ yields the largest improvement in overall mAP, indicating that a larger backbone is better able to recover discriminative cues under severe scale compression and low-resolution observations. Importantly, these gains are realized only when backbone scaling is coupled with stability-oriented selective fine-tuning, which helps preserve the pretrained feature space while enabling high-level domain adaptation.
Table \ref{tab:table5} and Figure \ref{fig:fig4} further show that the proposed method improves mAP across all three protocols, with the largest gains observed in the cross-view settings A$2$G and G$2$A. This suggests that the combination of selective fine-tuning, prompt-conditioned adaptation, and temporal attention is particularly effective under strong viewpoint mismatch and scale variation. In addition, re-ranking and temporal attention provide complementary improvements by improving neighborhood consistency at inference time and suppressing degraded frames within noisy tracklets, respectively.

\paragraph{\textit{Limitations.}}
Although our method achieves the best overall mAP and the best protocol-wise mAP across A$2$A, A$2$G, and G$2$A, it does not outperform prior methods on every ranking metric. In particular, the best public result remains stronger on some Rank-$5$ and Rank-$10$ comparisons, and also attains a higher Rank-$1$ on A$2$A. This suggests that while our approach improves top-ranked retrieval quality and overall ranking consistency, there is still room to improve deeper-list retrieval and same-view aerial matching. In addition, our evaluation is limited to the DetReIDX benchmark, and the current ablation analysis focuses on overall mAP, which restricts finer protocol-wise attribution of individual components. Finally, the larger ViT-L/$14$ backbone and optional re-ranking introduce additional computational cost and inference latency compared to the baseline.

\paragraph{\textit{Future work.}}
Several directions may further improve extreme far-distance ReID. First, protocol-wise ablations and retrieval calibration could provide deeper insight into which components benefit A$2$G, G$2$A, and A$2$A most, and may help improve deeper-rank retrieval behavior. Second, more expressive scale-aware prompting, metadata-conditioned adapters, or hierarchical temporal modeling may further improve robustness under severe altitude, distance, and motion variation. Third, evaluating cross-dataset generalization on additional aerial ReID benchmarks would provide stronger evidence of robustness beyond DetReIDX. Finally, future work should also explore more efficient adaptation strategies for large vision-language backbones and consider privacy-aware evaluation for aerial person recognition systems.

\section{Conclusion}
\label{sec:conclusion}

We present a scale-aware adaptation framework for CLIP-based video person re-identification in extreme far-distance aerial-ground scenarios. By scaling the visual backbone from ViT-B/$16$ to ViT-L/$14$ and combining it with stability-oriented selective fine-tuning, temporal attention pooling, optimization refinements, and k-reciprocal re-ranking, our method substantially improves retrieval robustness under severe scale compression and cross-view mismatch. On the DetReIDX benchmark, our approach achieves an overall mAP of $35.73$, with protocol-wise mAPs of $46.69$ on A$2$G, $41.23$ on G$2$A, and $22.98$ on A$2$A, outperforming both the official baseline and the best public result. These findings demonstrate that large-scale vision-language backbones, when carefully adapted, provide an effective foundation for extreme far-distance video person ReID.

\section*{Acknowledgments}

The authors acknowledge Research Computing at Arizona State University for providing HPC and storage resources \cite{jennewein2023sol} that contributed to the results reported in this work.

{
    \small
    \bibliographystyle{unsrt}
    \bibliography{main}

@article{wu2020reidsurvey,
  title   = {Deep Learning for Person Re-Identification: A Survey and Outlook},
  author  = {Wu, Lin and Wang, Yang and Gao, Jianhuang and Li, Xuelong},
  journal = {arXiv preprint arXiv:2001.04193},
  year    = {2020}
}

@article{radford2021clip,
  title   = {Learning Transferable Visual Models From Natural Language Supervision},
  author  = {Radford, Alec and Kim, Jong Wook and Hallacy, Chris and Ramesh, Aditya and Goh, Gabriel and Agarwal, Sandhini and Sastry, Girish and Askell, Amanda and Mishkin, Pamela and Clark, Jack and Krueger, Gretchen and Sutskever, Ilya},
  journal = {arXiv preprint arXiv:2103.00020},
  year    = {2021}
}

@inproceedings{zhong2017reranking,
  title     = {Re-Ranking Person Re-Identification with k-Reciprocal Encoding},
  author    = {Zhong, Zhun and Zheng, Liang and Cao, Donglin and Li, Shaozi},
  booktitle = {Proceedings of the IEEE Conference on Computer Vision and Pattern Recognition (CVPR)},
  year      = {2017}
}

@article{hambarde2025detreidx,
  title   = {{DetReIDX}: A Stress-Test Dataset for Real-World {UAV}-Based Person Recognition},
  author  = {Hambarde, Khushal and Mbongo, Emmanuel and Menghani, Nirav and Ramesh, Ajay and Feris, Rogerio Schmidt and Proen{\c{c}}a, Hugo},
  journal = {arXiv preprint arXiv:2505.04793},
  year    = {2025}
}

@inproceedings{zheng2015market,
  title     = {Scalable Person Re-identification: A Benchmark},
  author    = {Zheng, Liang and Shen, Liyue and Tian, Lu and Wang, Shengjin and Wang, Jingdong and Tian, Qi},
  booktitle = {Proceedings of the IEEE International Conference on Computer Vision (ICCV)},
  year      = {2015}
}

@inproceedings{zheng2017unlabeled,
  title     = {Unlabeled Samples Generated by {GAN} Improve the Person Re-identification Baseline in vitro},
  author    = {Zheng, Zhedong and Zheng, Liang and Yang, Yi},
  booktitle = {Proceedings of the IEEE International Conference on Computer Vision (ICCV)},
  year      = {2017}
}

@inproceedings{ristani2016dukemtmc,
  title     = {Performance Measures and a Data Set for Multi-Target, Multi-Camera Tracking},
  author    = {Ristani, Ergys and Solera, Francesco and Zou, Roger and Cucchiara, Rita and Tomasi, Carlo},
  booktitle = {European Conference on Computer Vision Workshops (ECCV Workshops)},
  year      = {2016}
}

@inproceedings{he2021transreid,
  title     = {TransReID: Transformer-Based Object Re-Identification},
  author    = {He, Shuting and Wu, Haoxi and Wang, Peng and Zhang, Mengdan and Huang, Zhongzhan and Tian, Yonghong},
  booktitle = {Proceedings of the IEEE/CVF International Conference on Computer Vision (ICCV)},
  year      = {2021}
}

@inproceedings{zheng2016mars,
  title     = {{MARS}: A Video Benchmark for Large-Scale Person Re-Identification},
  author    = {Zheng, Liang and Bie, Zhi and Sun, Yifan and Wang, Jingdong and Su, Chi and Wang, Shengjin and Tian, Qi},
  booktitle = {European Conference on Computer Vision (ECCV)},
  year      = {2016}
}

@inproceedings{fu2019sta,
  title     = {Spatial-Temporal Attention Model for Video-Based Person Re-Identification},
  author    = {Fu, Yang and Wang, Xiaoyang and Wei, Yunchao and Huang, Thomas},
  booktitle = {AAAI Conference on Artificial Intelligence (AAAI)},
  year      = {2019}
}

@inproceedings{hou2020tclnet,
  title     = {{TCLNet}: Temporal Complementary Learning for Video Person Re-identification},
  author    = {Hou, Ruibing and Chang, Hong and Ma, Bingpeng and Shan, Shiguang and Chen, Xilin},
  booktitle = {European Conference on Computer Vision (ECCV)},
  year      = {2020}
}

@inproceedings{li2021uavhuman,
  title     = {{UAV-Human}: A Large Benchmark for Human Behavior Understanding with Unmanned Aerial Vehicles},
  author    = {Li, Tianjiao and Liu, Jun and Zhang, Wei and Ni, Yun and Wang, Wenqian and Li, Zhiheng},
  booktitle = {Proceedings of the IEEE/CVF Conference on Computer Vision and Pattern Recognition (CVPR)},
  year      = {2021}
}

@article{kumar2021pdestre,
  title   = {The {P-DESTRE}: A Fully Annotated Dataset for Pedestrian Detection, Tracking, and Short/Long-Term Re-Identification from Aerial Devices},
  author  = {Kumar, S. V. Aruna and Yaghoubi, Ehsan and Das, Abhijit and Harish, B. S. and Proen{\c{c}}a, Hugo},
  journal = {IEEE Transactions on Information Forensics and Security (TIFS)},
  volume  = {16},
  pages   = {1696--1708},
  year    = {2021}
}

@inproceedings{nguyen2023aerialground,
  title     = {Aerial-Ground Person Re-Identification},
  author    = {Nguyen, Huy and Nguyen, Kien and Sridharan, Sridha and Fookes, Clinton},
  booktitle = {IEEE International Conference on Multimedia and Expo (ICME)},
  year      = {2023},
  pages     = {2585--2590}
}

@article{zhou2022cocoop,
  title   = {Conditional Prompt Learning for Vision-Language Models},
  author  = {Zhou, Kaiyang and Yang, Jingkang and Loy, Chen Change and Liu, Ziwei},
  journal = {arXiv preprint arXiv:2203.05557},
  year    = {2022}
}

@inproceedings{dosovitskiy2021vit,
  title     = {An Image is Worth 16x16 Words: Transformers for Image Recognition at Scale},
  author    = {Dosovitskiy, Alexey and Beyer, Lucas and Kolesnikov, Alexander and Weissenborn, Dirk and Zhai, Xiaohua and Unterthiner, Thomas and Dehghani, Mostafa and Minderer, Matthias and Heigold, Georg and Gelly, Sylvain and Uszkoreit, Jakob and Houlsby, Neil},
  booktitle = {International Conference on Learning Representations (ICLR)},
  year      = {2021}
}

@article{hermans2017triplet,
  title   = {In Defense of the Triplet Loss for Person Re-Identification},
  author  = {Hermans, Alexander and Beyer, Lucas and Leibe, Bastian},
  journal = {arXiv preprint arXiv:1703.07737},
  year    = {2017}
}

@inproceedings{loshchilov2017sgdr,
  title     = {{SGDR}: Stochastic Gradient Descent with Warm Restarts},
  author    = {Loshchilov, Ilya and Hutter, Frank},
  booktitle = {International Conference on Learning Representations (ICLR)},
  year      = {2017}
}

@inproceedings{houlsby2019adapters,
  title     = {Parameter-Efficient Transfer Learning for {NLP}},
  author    = {Houlsby, Neil and Giurgiu, Andrei and Jastrzebski, Stanislaw and Morrone, Bruna and de Laroussilhe, Quentin and Gesmundo, Andrea and Attariyan, Mohammad and Gelly, Sylvain},
  booktitle = {International Conference on Machine Learning (ICML)},
  year      = {2019}
}

@inproceedings{hu2022lora,
  title     = {{LoRA}: Low-Rank Adaptation of Large Language Models},
  author    = {Hu, Edward and Shen, Yelong and Wallis, Phillip and Allen-Zhu, Zeyuan and Li, Yuanzhi and Wang, Shean and Wang, Lu and Chen, Weizhu},
  booktitle = {International Conference on Learning Representations (ICLR)},
  year      = {2022}
}

@inproceedings{bai2022salient,
  author       = {Shutao Bai and Bingpeng Ma and Hong Chang and Rui Huang and Xilin Chen},
  title        = {Salient-to-Broad Transition for Video Person Re-Identification},
  booktitle    = {Proceedings of the IEEE/CVF Conference on Computer Vision and Pattern Recognition (CVPR)},
  pages        = {7339--7348},
  year         = {2022}
}

@inproceedings{hou2021bicnet,
  author       = {Ruibing Hou and Hong Chang and Bingpeng Ma and Rui Huang and Shiguang Shan},
  title        = {BiCnet-TKS: Learning Efficient Spatial-Temporal Representation for Video Person Re-Identification},
  booktitle    = {Proceedings of the IEEE/CVF Conference on Computer Vision and Pattern Recognition (CVPR)},
  pages        = {2014--2023},
  year         = {2021}
}

@inproceedings{wang2021pyramid,
  author       = {Yingquan Wang and Pingping Zhang and Shang Gao and Xia Geng and Hu Lu and Dong Wang},
  title        = {Pyramid Spatial-Temporal Aggregation for Video-Based Person Re-Identification},
  booktitle    = {Proceedings of the IEEE/CVF International Conference on Computer Vision (ICCV)},
  pages        = {12026--12035},
  year         = {2021}
}

@inproceedings{zhang2024crossplatform,
  author       = {Shizhou Zhang and Wenlong Luo and De Cheng and Qingchun Yang and Lingyan Ran and Yinghui Xing and Yanning Zhang},
  title        = {Cross-Platform Video Person ReID: A New Benchmark Dataset and Adaptation Approach},
  booktitle    = {European Conference on Computer Vision (ECCV)},
  pages        = {270--287},
  publisher    = {Springer},
  year         = {2024}
}

@misc{hambarde2026vreidxfd,
  title        = {{VReID-XFD: Video-based Person Re-identification at Extreme Far Distance Challenge Results}},
  author       = {Kailash A. Hambarde and Hugo Proença and Md Rashidunnabi and Pranita Samale and Qiwei Yang and Pingping Zhang and Zijing Gong and Yuhao Wang and Xi Zhang and Ruoshui Qu and Qiaoyun He and Yuhang Zhang and Thi Ngoc Ha Nguyen and Tien-Dung Mai and Cheng-Jun Kang and Yu-Fan Lin and Jin-Hui Jiang and Chih-Chung Hsu and Tamás Endrei and György Cserey and Ashwat Rajbhandari},
  year         = {2026},
  eprint       = {2601.01312},
  archivePrefix= {arXiv},
  primaryClass = {cs.CV},
  doi          = {10.48550/arXiv.2601.01312},
  url          = {https://arxiv.org/abs/2601.01312},
  note         = {Submitted on 4 Jan 2026}
}

@incollection{jennewein2023sol,
  title={The sol supercomputer at arizona state university},
  author={Jennewein, Douglas M and Lee, Johnathan and Kurtz, Chris and Dizon, William and Shaeffer, Ian and Chapman, Alan and Chiquete, Alejandro and Burks, Josh and Carlson, Amber and Mason, Natalie and others},
  booktitle={Practice and Experience in Advanced Research Computing 2023: Computing for the Common Good},
  pages={296--301},
  year={2023}
}
}


\end{document}